# Bioclimatic Modelling: A Machine Learning Perspective

Maumita Bhattacharya
School of Computing & Mathematics
Charles Sturt University
Albury, Australia
Maumita.bhattacharya@ieee.org

*Abstract*—Many machine learning (ML) approaches are widely used to generate bioclimatic models for prediction of geographic range of organism as a function of climate. Applications such as prediction of range shift in organism, range of invasive species influenced by climate change are important parameters in understanding the impact of climate change. However, success of machine learning-based approaches depends on a number of factors. While it can be safely said that no particular ML technique can be effective in all applications and success of a technique is predominantly dependent on the application or the type of the problem, it is useful to understand their behaviour to ensure informed choice of techniques. This paper presents a comprehensive review of machine learning-based bioclimatic model generation and analyses the factors influencing success of such models. Considering the wide use of statistical techniques, in our discussion we also include conventional statistical techniques used in bioclimatic modelling.

## I. Introduction

Understanding species' geographic range has become all the more important with concerns over global climatic changes and possible consequential range shifts, spread of invasive species and impact on endangered species. The key methods used to study geographic range are bioclimatic models, alternatively known as envelope models [38], climate response surface models [34], ecological niche models [66] or species distribution models [41]. Predictive ability lies at the core of such methods as it is the ultimate goal of ecology [63].

Machine Learning (ML) as a research discipline has roots in Artificial Intelligence and Statistics and the ML techniques focus on extracting knowledge from datasets [51]. This knowledge is represented in the form of a model which provides description of the given data and allows predictions for new data. This predictive ability makes ML a worthy candidate for bioclimatic modelling. Many ML algorithms are showing promising results in bioclimatic modelling including modelling and prediction of species distribution [20]. There are diverse applications of ML algorithms in ecology. They range from experimenting bio-geographical, ecological, and also evolutionary hypotheses to modelling species distributions for conservation, management and future planning [22, 71, 13, 21, and 59]. In the context of Eco-informatics [27] machine learning (ML) is a fast growing area which is concerned with finding patterns in complex, often nonlinear and noisy data and generating predictive models of relatively high accuracy. The increase in use of the ML techniques in ecological modelling in recent years is justified by the fact that this ability to produce predictive models of high accuracy does not involve the restrictive assumptions required by conventional, parametric approaches [29, 66, 54 and 20].

It may be noted that there is no universally best ML method; choice of a particular method or a combination of such methods is largely dependent on the particular application and requires human intervention to decide about the suitability of a method. However, concrete understanding of their behavior while applied to bioclimatic modelling can assist selection of appropriate ML technique for specific bioclimatic modelling applications.

In this paper we present a concise review of application of machine learning approaches to bioclimatic modelling and attempt to identify the factors that influence success or failure of such applications. In our discussion we have also included popular applications of statistical techniques to bioclimatic modelling.

The rest of the paper is organized as follows: Section II provides an overview of the Machine Learning and statistical methods commonly used in bioclimatic modelling and their applications to bioclimatic modelling; Section III presents an investigation on factors which influence success of such applications; finally in Section IV, we present some concluding remarks.

## II. ML & Statistical Techniques & Their Application to Bioclimatic Modelling

The inference mechanisms employed by Machine Learning (ML) techniques involve drawing conclusions from a set of examples. Supervised learning is one of the key ML inference mechanisms and is of particular interest in prediction of geographic ranges. In supervised learning the information about the problem being modeled is presented by datasets comprising input and desired output pairs [51]. The ML inference mechanism extracts knowledge representation from these examples to predict outputs for new inputs. The ML inference mechanism is depicted in Fig. 1.

The relatively more popular bioclimatic modelling applications of statistical and machine learning techniques and features of the relevant techniques are discussed next.

## A. Statistical Approaches

### 1) Generalised Linear Model (GLM)

Generalised linear models (GLM) [46] are probably the most commonly used statistical methods in the bioclimatic modelling community and have proven ability to predict current species distribution [7].

Generalised linear model (GLM) is a flexible generalization of regular linear regression. In GLM the response variable is normally modeled as a linear function of the independent variables. The degree of the variance of each measurement is a function of its predicted value.

Logistic regression analysis has been widely used in many disciplines including medical, social and biological sciences [33]. Its bioclimatic modelling application is relatively straightforward where a binary response variable is regressed against a set of climate variables as independent variables.

### 2) Generalised Additive Model (GAM)

Considering the limitations of Generalised Linear Models in capturing complex response curves, application of Generalised Additive Models is being proposed for species suitability modelling [6, 78 and 5].

The Generalised Additive Model (GAM) blends the properties of the Generalised Linear Models and Additive models [25]. GAM is based on non-parametric regression and unlike GLM does not impose the assumption that the data supports a particular functional form (normally linear). Here the response variable is the additive combination of the independent variables' functions. However, transparency and interpretability are compromised to accommodate this greater flexibility.

GAM can be used to estimate a non-constant species' response function, where the function depends on the location of the independent variables in the environmental space.

### 3) Climate Envelope Techniques

There are a number of specialized statistics-based tools developed for the purpose of bioclimatic modelling. Climate envelope techniques such as ANUCLAM, BIOCLIM, DOMAIN, FEM and HABITAT are popular and specialized bioclimatic modelling tools and thus deserve mention here. These tools usually fit a minimal envelope in a multidimensional climate space. Also, they use presence-only data instead of presence-absence data. This is highly beneficial as many data sets contain presence-only data.

Other statistical methods gaining popularity includes the Multivariate Adaptive Regression Splines [19].

## B. Machine Learning Approaches

### 1) Evolutionary Algorithms (EA)

Evolutionary Algorithms are basically stochastic and iterative optimisation techniques with metaphor in natural evolution and biological sexual reproduction [32 and 26]. Over the years several algorithms have been developed which fall in this category; some of the more popular ones being Genetic Algorithm, Evolutionary Programming, Genetic Programming, Evolution Strategy, Differential Evolution and so on. The most popular and extensive application of Evolutionary Algorithm and more specifically Genetic Algorithm (GA) to bioclimatic modelling has been through the software Genetic Algorithm for Rule-set Production (GARP) [4, 66 and 67]. Here, we shall restrict our discussion on application of Evolutionary Algorithm to bioclimatic modelling primarily to GARP.

Genetic Algorithm for Rule Set Production [81] is a specialised software based on Genetic Algorithm [51] for ecological niche modelling. The GARP model is represented by a set of mathematical rules based on environmental conditions. Each set of rules is an individual in the GA population. These rule sets are evolved through GA iterations. The model predicts presence of a species if all rules are satisfied for a specific environmental condition. The four sets of rules which are possible are: atomic, logistic regression, bio-climatic envelope and negated bio-climatic envelope [42].

GARP is essentially a non-deterministic approach that produces Boolean responses (presence/absence) for each environmental condition. As in case of the climate envelope techniques, GARP also does not require presence/absence data and can handle presence-only data. However, as the "learning" in GARP is based on optimisation of a combination of several types of models and not of one particular type of model, ecological interpretability may be difficult.

Examples of applications of GARP for bioclimatic modelling include: the habitat suitability modelling of threatened species [3] and that of invasive species [66, 65 and 17], and the geography of disease transmission [64].

Other applications of Genetic Algorithm to ecological modelling include: modelling of the distribution of cutthroat and rainbow trout as a function of stream habitat characteristics in the Pacific Northwest of the USA [14] and modelling of plant species distributions as a function of both climate and land use variables [82]. McKay [47] used Genetic Programming (GP) to develop spatial models for marsupial density. Chen et al. [11] used GP to analyse fish stock-recruitment relationship, and Muttil and Lee [52] used this technique to model nuisance algal blooms in coastal ecosystems. Newer approaches to use Evolutionary Algorithms for ecological niche modelling are being proposed such as the WhyWhere algorithm advocated by Stockwell [80]. EC has also been applied in conservation planning for biodiversity [75].

### 2) Artificial Neural Network (ANN)

A relatively later introduction to species distribution modelling is that of the Artificial Neural Network (ANN) [43, 57, 67 and 83].

TABLE I. COMPARISON OF SOME OF THE RELEVANT CHARACTERISTICS OF ML TECHNIQUES

| Characteristic | GLM | DT | ANN | EA |
|---|---|---|---|---|
| Mixed data handling ability | Low | High | Low | Moderate |
| Outlier handling ability | Low | Moderate | Moderate | Moderate |
| Non-linear relationship modelling | Low | Moderate | High | High |
| Transparency of modelling process | High | Moderate | Low | Low |
| Predictive ability | Low | Moderate | High | High |

Artificial Neural Networks are computational techniques with metaphor in the structure, processing mechanism and learning ability of the brain [30]. The processing units in ANN simulate biological neurons and are known as nodes. These artificial neurons or nodes are organised in one or more layers. Simulating the biological synapses, each node is connected to one or more nodes through weighted connections. These weights are adjusted to acquire and store knowledge about data. There are many algorithms available to train the ANN.

Some of the noteworthy applications of ANN are as follows: species distribution modelling [45 and 53], species diversity modelling [28, 10 and 56], community composition modelling [55], aquatic primary and secondary production modelling [76 and 48], species classification in appropriate taxonomic groups using multi-locus genotypes [12], modelling of wildlife damage to farmlands [79], assessment of potential impacts of climate change on distribution of tree species in Europe [83], invasive species modelling [87], and pest management [90]. Please see Olden et al. [58] for further details.

The main advantages of ANNs are that they are robust, perform well with noisy data and can represent both linear and non-linear functions of different forms and complexity levels. Their ability to handle non-linear responses to environmental variables is an advantage. However, they are less transparent and difficult to interpret. Inability to identify the relative importance and effect of the individual environmental variables is a limitation [83].

*3) Decision Trees (DT)*

Decision Trees have also found numerous applications in bioclimatic modelling. Decision Trees represent the knowledge extracted from data in a recursive, hierarchical structure comprising of nodes and branches [69]. Each internal node represents an input variable or attribute. They are associated with a test or decision rule relevant to data classification. Each leaf node represents a classification or a decision i.e. the value of the target variable conditional to the value of the input variables represented by the root to leaf path. Predictions derived from a Decision Tree generally involve determination of a series of 'if-then-else' conditions [8].

The two main types of Decision Trees used for predictions are: Classification Tree analysis and Regression Tree Analysis. The term Classification and Regression Tree (CART) analysis is the umbrella term used to refer to both Classification Tree analysis and Regression Tree analysis [8].

Some of the relevant and relatively recent applications of Decision Trees are as follows: habitat models for tortoise species [2], and endangered crayfishes [86]; quantification of the relationship between frequency and severity of forest fires and landscape structure by Rollins et al. [73]; prediction of fish species invasions in the Laurentian Great Lakes by Mercado-Silva et al. [49]; species distribution modelling of bottlenose dolphin [84]; development of models to assess the vulnerability of the landscape to tsunami damage [35]. Olden et al. [58] provides a more complete list.

The obvious advantage of the Decision Trees is that the ecological interpretability of the results derived from them is simple. Also there is no assumed functional relationship between the environmental variables and species suitability [15, 74 and 89]. Despite their ease of interpretability, Decision Trees may suffer from over-fitting [8 and 83].

Some relevant characteristics of different ML techniques are depicted in Table 1. Also see [58].

### III. FACTORS INFLUENCING SUCCESS OF ML APPROACHES TO BIOCLIMATIC MODELLING

While it is not that straightforward to identify the causes of success or failure of applications of the Machine Learning techniques to bioclimatic modeling, in this section we attempt to outline some of the factors which may impact their performance broadly. However, this is not to undermine the fact that success or failure of any machine learning application is predominantly dependent on the specific application.

*A. Very large data sets*

Data sets with hundreds of fields and tables and millions of records are commonplace and may pose challenge to the ML processors. However, enhanced algorithms, effective sampling, approximation and massively parallel processing offer solution to this problem.

*B. High dimensionality*

Many bioclimatic modeling problems may require a large number of attributes to define the problem. Machine learning algorithms struggle when they are to deal with not just large data sets with millions of records, but with a large number of fields or attributes, increasing the dimensionality of the problem. A high dimensional data set poses challenges by increasing the search space for model induction. This also increases the chances of the ML algorithm finding invalid patterns. Solution to this problem includes reducing dimensionality and using prior knowledge to identify irrelevant attributes.

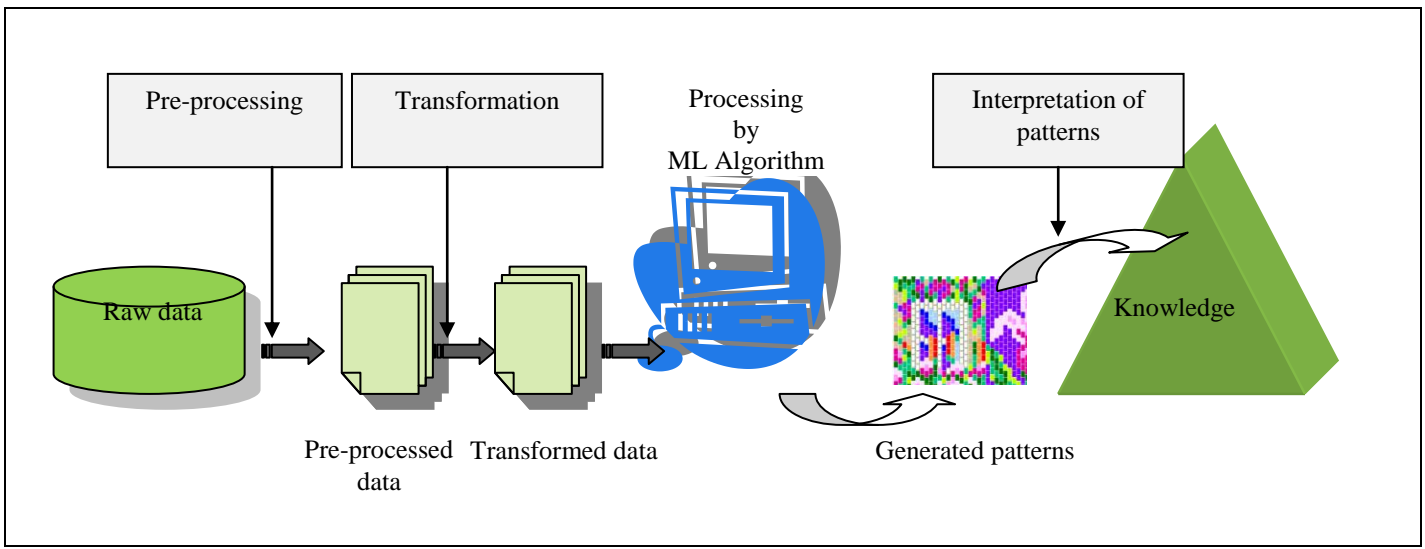

Figure 1. Steps involved in the Machine Learning inference process

### C. Over-fitting

Over-fitting occurs when the algorithm can model not only the valid patterns in the data but also any noise specific to the data set. This leads to poor performance as it can exaggerate minor fluctuations in the data. Decision Tress and also some of the Artificial Neural Networks may suffer from over-fitting. Cross-validation and regularization are some of the possible solutions to this problem.

### D. Dynamic environment

Rapidly changing or dynamic data makes it hard to discover patterns as previously discovered patterns may become invalid. Values of the defining variables may become unstable. Incremental methods that are capable of updating the patterns and identifying the patterns of changes hold the solution.

### E. Noisy and missing data

This problem is not uncommon in ecological data sets. Data smoothing techniques may be used for noisy data. Statistical strategies to identify hidden variables and dependencies may also be used.

### F. Complex dependencies among attributes

The traditional Machine Learning techniques are not necessarily geared to handle complex dependencies among the attributes. Techniques which are capable of deriving dependencies between variables have also been experimented in the context of data mining [18 and 16].

### G. Interpretability of the generated patterns

Ecological interpretability of the generated patterns is a major issue in many of the ML applications to bioclimatic modeling. Applications of Evolutionary Algorithm and Artificial Neural Networks may suffer from poor interpretability. Decision Trees on the other hand scores high in terms of interpretability.

Other influencing factors, which are not directly related to the characteristic of Machine Learning techniques, are as below.

### H. Choice of test and training data

Various reported applications of ML used the following three different means to choose test and training data: *re-substitution* – the same data set is used for both training and testing; *data splitting* – the data set is split into a training set and a test set; *independent validation* – the model is fitted with a data set independent of the test data set. Naturally, independent validation is the preferred method in most cases, followed by data splitting and then re-substitution. The results obtained by data splitting and re-substitution may be overly optimistic due to over-fitting [36]. However, the choice of one technique over the other is also problem dependent. Only a small segment of the reported studies seems to use independent validation.

### I. Model evaluation metrics

The measure of model performance or the model evaluation technique should ideally be chosen based on the purpose of the study or the modeling exercise. It is thus perfectly understandable that different authors have used different evaluation metric for their specific studies. Pease see the following literature for further discussions on choice of evaluation metrics: Fielding & Bell [23]; Guisan & Zimmermann [29]; Pearce & Ferrier [60]; Manel et al. [44]; Fielding [24]; Liu et al. [40]; Vaughan & Ormerod [88]; Allouche et al. [1].

TABLE II. FACTORS INFLUENCING APPLICATION OF ML TECHNIQUES TO ECOLOGICAL MODELLING

| Factor | Impact on ML technique | Possible solution |
|---|---|---|
| Very large data sets | EC, ANN and DT all are adversely effected | Enhanced algorithms, effective sampling, approximation; massively parallel processing |
| High dimensionality | EC, ANN and DT all are adversely effected | Reducing dimensionality; using prior knowledge to identify irrelevant attributes |
| Over-fitting | DT and some of the ANNs are adversely effected | Cross-validation; regularization |
| Dynamic environment | EC, ANN and DT all are adversely effected | Incremental methods capable of updating the patterns and identifying the patterns of changes |
| Noisy and missing data | DT is better equipped to handle this problem compared to others | Data smoothing; Statistical strategies to identify hidden variables and dependencies |
| Complex dependencies among attributes | EC, ANN and DT all are effected; however, handles better than traditional techniques such as GLM | |
| Interpretability of the generated patterns | EC = poor interpretability; DT and ANN= moderate to high interpretability | |
| Choice of test and training data | Effects EC, ANN and DT | Depends on goal of the study; however, generally independent validation is better than others |
| Model evaluation metrics | Effects EC, ANN and DT | Depends on goal of the study |

Table 2 summarises the factors influencing application of ML techniques to ecological modelling and Table 3 presents the comparative performances of some of the selected studies found in the literature [36].

As can be seen, none of the modeling techniques is universally superior compared to other techniques across all applications. Comparative performances of the three traditional methods, namely, GLM, GAM and climate envelope model, shows GAM and GLM have comparable performances. Among the Machine Learning methods, the popular GARP technique produces moderate performance, while CART and ANN have shown mixed results. It may be noted that these examples did not include adequate number of applications of ANN. Jeschke and Strayer [36] have reported, overall, ANN performs better among the ML techniques applied to this problem domain. Robustness is a characteristic often attributed to ANN. The findings by Jeschke and Strayer [36] also validate this claim. The specialized climate envelope techniques such as BIOCLIM, FEM and DOMAIN show only moderate performance in general and often perform worse than the Machine Learning techniques. However, some of the relatively recent comparisons have claimed that newer techniques are likely to outperform more established techniques (e.g. the model-averaging random forests by Lawler et al. [39] and Broennimann et al. [9]; the Bayesian weights-of-evidence model by Zeman & Lynen [91]). However, as these methods have been used only in a handful of studies, claims about their predictive power is premature [36]. Finally, this comparative study reiterates the fact that success and failure of the modelling techniques is primarily dependent on the application including the data set and the goal of study.

## IV. CONCLUSION

This paper presented a comprehensive review of applications of various Machine Learning techniques to bioclimatic modelling and broadly to ecological modelling. Some of the statistical techniques popular in this application domain have also been discussed. Factors influencing the performance of such techniques have been identified. Performances of these techniques when applied to ecological modeling have been compared based on studies reported in existing literature. It has been concluded that success or failure of application of the Machine Learning techniques to ecological modeling is primarily application dependent and none of techniques can claim superior performance as against other techniques universally. However, the identified factors or characteristics can be used as a guideline to select the ML techniques for such modeling exercises.

REFERENCES

[1] Allouche,O., A. Tsoar & R.Kadmon, Assessing the accuracy of species distribution models: prevalence, kappa and the true skill statistic (TSS), *J. Applied Ecology*, Vol.43, 2006, pp.1223–1232.

[2] Anderson M. C., Watts J. M., Freilich J. E., Yool S. R., Wakefield G. I., McCauley J. F., Fahnestock P. B., Regression-tree modeling of desert tortoise habitat in the central Mojave desert. *Ecological Applications*, Vol.10, No.3, 2000, pp.890–900.

[3] Anderson R. P., Martı́nez-Meyer E., Modeling species' geographic distributions for preliminary conservation assessments: an implementation with the spiny pocket mice (*Heteromys*) of Ecuador, *Biological Conservation*, Vol.116, No.2, 2004, pp.167–179.

[4] Anderson, R.P., Lew, D., Peterson, A.T., Evaluating predictive models of species' distributions: criteria for selecting optimal models, *Ecological. Modelling*, Vol.162, 2003, pp.211–232.

[5] Austin M., Species distribution models and ecological theory: a critical assessment and some possible new approaches, *Ecological Modelling*, Vol.200 (1–2): 2007, pp.1–19.

[6] Austin, M.P., Meyers, J.A., Current approaches to modelling the environmental niche of eucalyptus: implication for management of forest biodiversity, *Forest Ecology Management*, Vol.85, 1996, pp.95–106.

[7] Bakkenes, M., Alkemade, J.R.M., Ihle, R., Leemans, R., Latour, J.B., Assessing effects of forecasting climate change on the diversity and distribution of European higher plants for 2050, *Global Change Biology*, Vol.8, 2002, pp.390–407.

[8] Breiman, L., Friedman, J. H., Olshen, R. A., & Stone, C. J., Classification and regression trees, Monterey, CA: Wadsworth & Brooks/Cole Advanced Books & Software. ISBN 978-0412048418, 1984.

[9] Broennimann, O. et al., Evidence of climatic niche shift during biological invasion, *Ecology Letters*, Vol.10, 2007, pp.701–709.

[10] Brosse S., Lek S., Townsend C. R., Abundance, diversity, and structure of freshwater invertebrates and fish communities: an artificial neural network approach, *New Zealand Journal of Marine and Freshwater Research*, Vol.35, No.1, 2001, pp.135–145.

[11] Chen D. G., Hargreaves N. B., Ware D. M., Liu Y. A fuzzy logic model with genetic algorithm for analyzing fish stock-recruitment relationships, *Canadian Journal of Fisheries and Aquatic Science*, Vol.57, No.9, 2000, pp.1878–1887.

[12] Cornuet J. M., Aulagnier S., Lek S., Franck P., Solignac M., Classifying individuals among infraspecific taxa using microsatellite data and neural networks, *Comptes rendus de l'Acade´mie des sciences, Se´rie III, Sciences de la vie*, Vol.319, No.12, 1996, pp.1167–1177.

[13] Cushing J. B., Wilson T., Eco-informatics for 190 THE QUARTERLY REVIEW OF BIOLOGY Volume 83 decision makers advancing a research agenda, Data Integration in the Life Sciences: Second International Workshop, DILS 2005, San Diego, CA, USA, July 20–22, 2005, Proceedings, Lecture Notes in Computer Science, Volume 3615, edited by B. Luda¨scher and L. Raschid. Berlin (Germany): Springer-Verlag, 2005, pp.325–334.

[14] D'Angelo D. J., Howard L. M., Meyer J. L., Gregory S. V., Ashkenas L. R., Ecological uses for genetic algorithms: predicting fish distributions in complex physical habitats. *Canadian Journal of Fisheries and Aquatic Sciences*, Vol.52, 1995, pp.1893–1908.

[15] De'Ath, G., Fabricius, K.E., Classification and regression treers: a powerful yet simple technique for ecological data analysis. *Ecology*, Vol.81, 2000, pp.3178–3192.

[16] Djoko, S.; Cook, D.; and Holder, L. Analyzing the Benefits of Domain Knowledge in Substructure Discovery, *Proceedings of KDD-95: First International Conference on Knowledge Discovery and Data Mining*, Menlo Park, Calif.: American Association for Artificial Intelligence, 1995, pp.5–80.

[17] Drake J. M., Lodge D. M., Forecasting potential distributions of nonindigenous species with a genetic algorithm. *Fisheries*, Vol.31, 2006, pp.9–16.

[18] Dzeroski, S. 1996. Inductive Logic Programming for Knowledge Discovery in Databases, *Advances in Knowledge Discovery and Data Mining*, eds. U. Fayyad, G. Piatetsky-Shapiro, P. Smyth, and R. Uthurusamy, Menlo Park, Calif.: AAAI Press, 1996, pp.59–82.

[19] Elith J., Leathwick J., Predicting species distributions from museum and herbarium records using multiresponse models fitted with multivariate adaptive regression splines. *Diversity and Distributions*, Vol.13, No.3, 2007, pp.265–275.

[20] Elith, J., Graham, C. H., Anderson, R. P., Dudk, M., Ferrier, S., Guisan, A., et al., Novel methods improve prediction of species' distributions from occurrence data, *Ecography*, Vol.29, 2006, pp.129–151.

[21] Ferrier S., Guisan A., Spatial modelling of biodiversity at the community level. *Journal of Applied Ecology*, Vol.43, No.3, 2006, pp.393–404.

[22] Fielding A. H., editor. *Machine Learning Methods for Ecological Applications*. Boston (MA): Kluwer Academic Publishers, 1999.

[23] Fielding, A.H. & J.F. Bell., A review of methods for the assessment of prediction errors in conservation presence/absence models. *Environmental Conservation*, Vol.24, 1997, pp.38–49.

[24] Fielding, A.H., What are the appropriate characteristics of an accuracy measurement? In *Predicting Species Occurrences: Issues of Accuracy and Scale*. J.M. Scott *et al.*, Eds., Island Press. Washington, D.C., 2002, pp.271–280.

[25] Friedman, J.H. and Stuetzle, W., Projection Pursuit Regression, *Journal of the American Statistical Association*, Vol.76, 1981, pp. 817–823.

[26] Goldberg D. E., Genetic Algorithms in Search, Optimization, and Machine Learning. Reading (MA): Addison-Wesley, 1989.

[27] Green, J. L., Hastings, A., Arzberger, P., Ayala, F. J., Cottingham, K. L., Cuddington, K., Davis, F., Dunne, J. A., Fortin M.J., Gerber, L., Neubert, M., Complexity in ecology and conservation: mathematical, statistical, and computational challenges, *BioScience*, Vol.55, No.6, 2005, pp.01–510.

[28] Gue´gan J.-F., Lek S., Oberdorff T., Energy availability and habitat heterogeneity predict global riverine fish diversity. *Nature*, Vol.391, 1998, pp.382–384.

[29] Guisan A., Zimmermann N. E., Predictive habitat distribution models in ecology, *Ecological Modelling*, Vol.135 (2–3), 2000, pp.147–186.

[30] Haykin, S., *Neural networks: A comprehensive foundation* (2nd ed.). Prentice Hall, 1998.

[31] Hernandez, P.A. et al., The effect of sample size and species characteristics on performance of different species distribution modelingmethods. *Ecography*, Vol.29, 2006, pp. 773–785.

[32] Holland J. H., Adaptation in Natural and Artificial Systems: An Introductory Analysis with Applications to Biology, Control, and Artificial Intelligence, Ann Arbor (MI): University of Michigan Press, 1975.

[33] Hosmer, D.W. and Lemeshow, S., *Applied Logistic Regression*, 2nd edn. Wiley. New York, 2000.

[34] Huntley, B., Plant-species response to climate change—implications for the conservation of European birds, *Ibis*, Vol.137, 1995, pp.S127—S138.

[35] Iverson L. R., Prasad A. M., Using landscape analysis to assess and model tsunami damage in Aceh province, Sumatra. *Landscape Ecology*, Vol.22, No.3, 2007, pp.323–331.

[36] Jeschke, J.M. and Strayer, D.L., Usefulness of Bioclimatic Models for Studying Climate Change and Invasive Species, *Annals of the New York Academy of Sciences*, Vol.1134, 2008, pp.1–24.

[37] Johnson, C.J. and Gillingham, M.P., An evaluation of mapped species distribution models used for conservation planning, *Environ. Conserv.* Vol.32, 2005, pp.117–128.

[38] Kadmon, R., Farber, O., and Danin, A., 2003. A systematicanalysis of factors affecting the performance ofclimatic envelope models. *Ecol. Appl.* Vol.13, 2003, pp.853–867.

[39] Lawler, J.J. *et al.* 2006. Predicting climate-induced range shifts: model differences and model reliability, *Global Change Biology*, Vol.12, 2006, pp.1568–1584.

[40] Liu, C. et al., Selecting thresholds of occurrence in the prediction of species distributions. *Ecography*, Vol.28, 2005, pp.385–393.

[41] Loiselle, B.A. et al., Avoiding pitfalls of using species distribution models in conservation planning, *Conservation. Biology*, Vol.17, 2003, pp.1591–1600.

[42] Lorena, A.C., Jacintho, L.F.O., Siqueira, M.F., Giovanni, R.D., Lohmann, L.G., Carvalho, A.C.P.L.F., and Yamamoto, M., Comparing Machine Learning Classifiers in Potential Distribution Modeling, *Expert Systems with Applications*, Vol. 38, 2011, pp.5268–5275.

[43] Manel, S., Dias, J.-M., Ormerod, S.J., Comparing discriminant analysis, neural networks and logistic regression for predicting species distributions: a case study with a Himalayan river bird, *Ecology Modellig*, Vol.120, 1999, pp.337–347.

[44] Manel, S., Williams, H.C. & Ormerod, S.J., Evaluating presence-absence models in ecology: the need to account for prevalence, *Journal of Applied Ecology*, Vol.38, 2001, pp.921–931.

[45] Mastrorillo S., Lek S., Dauba F., Belaud A., The use of artificial neural networks to predict the presence of small-bodied fish in a river, *Freshwater Biology*, Vol.38, No.2, 1997, pp.237–246.

[46] McCullagh, P., Nelder, J.A., *Generalized Linear Models*, Chapman & Hall, London, 1989.

[47] McKay R. I., Variants of genetic programming for species distribution modelling—fitness sharing, partial functions, population evaluation, *Ecological Modelling*, Vol.146 (1–3), 2001, pp.231–241.

[48] McKenna J. E., Jr., Application of neural networks to prediction of fish diversity and salmonid production in the Lake Ontario basin, *Transactions of the American Fisheries Society*, Vol.134, No.1, 2005, pp.28–43.

[49] Mercado-Silva N., Olden J. D., Maxted J. T., Hrabik T. R., Vander Zanden M. J., Forecasting the spread of invasive rainbow smelt in the Laurentian Great Lakes region of North America, *Conservation Biology*, Vol.20, No.6, 2006, pp.1740–1749.

[50] Meynard, C.N. and Quinn, J.F., Predicting species distributions: a critical comparison of the most common statistical models using artificial species, *Journal of Biogeography*, Vol.34, 2007, pp.1455–1469.

[51] Mitchell, T., *Machine learning*. McGraw Hill, 1997.
[52] Muttil N., Lee J. H. W., Genetic programming for analysis and real-time prediction of coastal algal blooms, *Ecological Modelling*, Vol.189(3–4), 2005, pp.363–376.
[53] Özesmi S. L., Özesmi U. 1999. An artificial neural network approach to spatial habitat modelling with interspecific interaction, *Ecological Modelling*, Vol.116, No.1, 1999, pp.15–31.
[54] Olden J. D., Jackson D. A., A comparison of statistical approaches for modelling fish species distributions, *Freshwater Biology*, Vol.47, No.10, 2002, pp.1976–1995.
[55] Olden J. D., Joy M. K., Death R. G., Rediscovering the species in community-wide modelling, *Ecological Applications*, Vol.16, No.4, 2006, pp.1449–1460.
[56] Olden J. D., Poff N. L., Bledsoe B. P., Incorporating ecological knowledge into ecoinformatics: an example of modeling hierarchically structured aquatic communities with neural networks, *Ecological Informatics*, Vol.1, No.1, 2006, pp.33–42.
[57] Olden, J.D., Jackson, D.A., Peres-Neto, P.R., Predictive models of fish species distributions: a note on proper validation and chance prediction, *Transactions of the American Fisheries Society*, Vol.131, 2002, pp.329–336.
[58] Olden, J.D., Lawler, J.J., Poff, N.L., Machine Learning Methods without Tears: A Primer for Ecologists, *The Quarterly Review of Biology*, Vol.83, No.2, 2008, pp.171–193.
[59] Park Y.-S., Chon T.-S., Biologically-inspired machine learning implemented to ecological informatics, *Ecological Modelling*, Vol.203 (1–2), 2007, pp.1–7.
[60] Pearce, J. and Ferrier, S., Evaluating the predictive performance of habitat models developed using logistic regression. *Ecology Modeling*, Vol.133, 2000, pp.225–245.
[61] Pearson, R.G. et al., Model-based uncertainty in species range prediction, *Journal of Biogeography*, Vol.33, 2006, pp.1704–1711.
[62] Pearson, R.G., Dawson, T.P., Berry, P.M., SPECIES: a spatial evaluation of climate impact on the envelope of species, *Ecological Modelling*, Vol.154, 2006, pp.289–300.
[63] Peters R. H., *A Critique for Ecology*. Cambridge (UK): Cambridge University Press, 1991.
[64] Peterson A. T., Predicting species' geographic distributions based on ecological niche modelling, *Condor*, Vol.103, No.3, 2001, pp.599–605.
[65] Peterson A. T., Predicting the geography of species' invasions via ecological niche modelling, *Quarterly Review of Biology*, Vol.78, No.4, 2003, pp.419–433.
[66] Peterson A. T., Vieglais D. A., Predicting species invasions using ecological niche modeling: new approaches from bioinformatics attack a pressing problem, *BioScience*, Vol.51, No.5, 2001, pp.363–371.
[67] Peterson, A.T., Ortega-Huerta, M.A., Bartley, J., Sanchez-Cordero, V., Soberon, J., Buddemeier, R.H., Stockwell, D.R.B., Future projections for Mexican faunas under global climate change scenarios, *Nature*, Vol.416, 2002, pp.626–629.
[68] Peterson, A.T., Sanchez-Cordero, V., Soberon, J., Bartley, J., Buddemeier, R.W., Navarro-Siguenza, A.G., Effects of global climate change on geographic distributions of Mexican Cracidae, *Ecological Modelling*, Vol.144, 2001, pp.21–30.
[69] Quinlan, J. R., Induction of decision trees. *Machine Learning*, Vol.1, No.1, 1986, pp.81–106.
[70] Randin, C.F. et al., Are niche-based species distribution models transferable in space?, *J.ournal of Biogeography*, Vol.33, 2006, pp.1689–1703.
[71] Recknagel F., Applications of machine learning to ecological modelling, *Ecological Modelling*, Vol.146 (1–3), 2001, pp.303–310.
[72] Robertson, M.P., Villet, M.H. & Palmer, A.R., A fuzzy classification technique for predicting species' distributions: applications using invasive alien plants and indigenous insects, *Diversity Distribution*, Vol.10, 2004, pp.461–474.
[73] Rollins M. G., Keane R. E., Parsons R. A., Mapping fuels and fire regimes using sensing, ecosystem simulation, and gradient modelling, *Ecological Applications*, Vol.14, No.1, 2004, pp.75–95.
[74] Rouget, M., Richardson, D.M., Milton, S.J., Polakow, D., Predicting invasion dynamics of four alien Pinus species in a highly fragmented semi-arid shrubland in South-Africa, *Plant Ecology*, Vol.152, 2001, pp.79–92.
[75] Sarkar S., Pressey R. L., Faith D. P., Margules C. R., Fuller T., Stoms D. M., Moffett A., Wilson K. A., Williams K. J., Williams P. H., Andelman S, Biodiversity conservation planning tools: present status and challenges for the future, *Annual Review of Environment and Resources*, Vol.31, 2006, pp.123–159.
[76] Scardi M., Harding L. W., Jr., Developing an empirical model of phytoplankton primary production: a neural network case study, *Ecological Modelling*, Vol.120 (2–3), 1999, pp.213–223.
[77] Schussman, H. et al., Spread and current potential distribution of an alien grass, Eragrostis lehmanniana Nees, in the southwestern USA: comparing historical data and ecological niche models, *Diversity Distribution*, Vol.12, 2006, pp.81–89.
[78] Seoane, J., Bustamante, J., Diaz-Delgado, R., Competing roles for landscape, vegetation, topography and climate in predictive models of bird distribution. *Ecological Modelling*, Vol.171, 2004, pp.209–222.
[79] Spitz F., Lek S., Environmental impact prediction using neural network modelling: an example in wildlife damage, *Journal of Applied Ecology*, Vol.36, No.2, 1999, pp.317–326.
[80] Stockwell D. R. B., Improving ecological niche models by data mining large environmental datasets for surrogate models, *Ecological Modelling*, Vol.192 (1–2), 2006, pp.188–196.
[81] Stockwell, D. R. B., & Peters, D. P., The GARP modelling system: Problems and solutions to automated spatial prediction, International Journal of Geographic Information Systems, Vol.13, 1999, pp.143–158.
[82] Termansen M., McClean C. J., Preston C. D., The use of genetic algorithms and Bayesian classification to model species distributions, *Ecological Modelling*, Vol.192 (3–4), 2006, pp.410–424.
[83] Thuiller, W.,BIOMOD - optimising predictions of species distributions and projecting potential future shifts under global change, *Global Change Biology*, Vol.9, 2003, pp.1353–1362.
[84] Torres L. G., Rosel P. E., D'Agrosa C., Read A. J., Improving management of overlapping bottlenose dolphin ecotypes through spatial analysis and genetics, *Marine Mammal Science*, Vol.19, No.3, 2003, pp.502–514.
[85] Tsoar, A. et al., A comparative evaluation of presence-only methods for modelling species distribution, *Diversity Distribution*, Vol.13, 2007, pp.397–405.
[86] Usio N., Endangered crayfish in northern Japan: distribution, abundance and microhabitat specificity in relation to stream and riparian environment, *Biological Conservation*, Vol.134, No.4, 2007, pp.517–526.
[87] Vander Zanden M. J., Olden J. D., Thorne J. H., Mandrak N. E., Predicting occurrences and impacts of smallmouth bass introductions in north temperate lakes, *Ecological Applications*, Vol.14, No.1, 2004, pp.132–148.
[88] Vaughan, I.P. & Ormerod, S.J., The continuing challenges of testing species distribution models. *Journal of Applied Ecology*, Vol.42, 2005, pp.720–730.
[89] Vayssieres, M.P., Plant, R.E., Allen-Diaz, B.H., Classification trees: an alternative non-parametric approach for predicting speciers distributions, *Journal of Vegetation Science*, Vol.11, 2000, pp.679–694.
[90] Worner S. P., Gevrey M., Modelling global insect pest species assemblages to determine risk of invasion, *Journal of Applied Ecology*, Vol.43, No.5, 2006, pp.858–867.
[91] Zeman, P. & Lynen, G., Evaluation of four modelling techniques to predict the potential distribution of ticks using indigenous cattle infestations as calibration data. *Experimental and Applied Acarology*, Vol.39,2006,pp.163–176.

TABLE III. COMPARATIVE PERFORMANCES OF ECOLOGICAL MODELLING TECHNIQUES**

| Study | Species | No of Unique Members of Species | Performance Evaluation Metric | Comparative Performance |
|---|---|---|---|---|
| Manel et al. [43] | Birds | 6 | Proportion of correct predictions, sensitivity, specificity, kappa and so on. | ANN>GLM |
| Termansen et al. [82] | Plants | 100 | AUC* score, sensitivity | GLM>GAM>CART |
| Vayssieres et al. [89] | Oaks | 3 | Sensitivity, specificity, differential positive rate | CART>GLM |
| Lorena et al. [42] | Plants | 35 | AUC score | SVM^>ANN>DT>GARP |
| Robertson et al. [72] | Plants | 3 | Kappa | FEM>BIOCLIM |
| Johnson and Gillingham [37] | *Rangifer tarandus caribou* | NA | r, $r_s$ | GLM>GARP |
| Elith et al. [20] | Animals, plants | 226 | AUC score, correlation, kappa | GAM>GLM≈BIOCLIM |
| Hernandez et al. [31] | Animal | 18 | AUC, sensitivity, kappa | GARP>BIOCLIM |
| Lawler et al. [39] | Mammal | 100 | AUC, sensitivity, kappa | Random forest>GLM>GAM≈ANN>CART≈GARP |
| Pearson et al. [62] | Proteaceae | 4 | AUC, kappa | GAM≈ANN>GLM>CART>GA>GARP>BIOCLIM |
| Schussman et al. [77] | Eragrostis lehmanniana | NA | Sensitivity, specificity, kappa | GLM>GARP |
| Randin et al. [70] | Plants | 54 | AUC score, kappa | GAM≈GLM |
| Zeman and Lynen [91] | Rhipicephalus appendiculatus | NA | Mean squared difference | Weights of evidence (Bayesian)>GAM |
| Meyard and Quinn [50] | Artificial species | 18 | AUC score, sensitivity, specificity, kappa | GAM>GLM>CART>GARP |
| Tsoar et al. [85] | Animals | 42 | Kappa | GARP>HABITAT>BIOCLIM |

\* Area under curve

^ Support vector machine

\*\* All experiments used data splitting as the means to select training and test sets.